\documentclass[twoside]{article}
\usepackage[accepted]{aistats2017}
\usepackage{comment}
\usepackage[round]{natbib}
\usepackage{amsmath,amsthm,amssymb,amsfonts}

\usepackage{algorithm}
\usepackage{algorithmic}
\usepackage{enumerate}
\usepackage{subfigure}
\usepackage{graphicx}
\usepackage{hyperref}
\usepackage{bm}
\usepackage[export]{adjustbox}

\DeclareMathOperator*{\argmax}{argmax}
\DeclareMathOperator*{\argmin}{argmin}

\newtheorem{theorem}{Theorem}
\begin{document}

%

%

\twocolumn[

\aistatstitle{Initialization and Coordinate Optimization for Multi-way Matching}

\aistatsauthor{ Da Tang \And Tony Jebara }

\aistatsaddress{ Columbia University \And  Columbia University $\&$ Netflix Inc.} ]

\begin{abstract}
We consider the problem of consistently matching multiple sets of elements to each other, which is a common task in fields such as computer vision. To solve the underlying NP-hard objective, existing methods often relax or approximate it, but end up with unsatisfying empirical performance due to a misaligned objective. We propose a coordinate update algorithm that directly optimizes the target objective. By using pairwise alignment information to build an undirected graph and initializing the permutation matrices along the edges of its Maximum Spanning Tree, our algorithm successfully avoids bad local optima. Theoretically, with high probability our algorithm guarantees an optimal solution under reasonable noise assumptions. Empirically, our algorithm consistently and significantly outperforms existing methods on several benchmark tasks on real datasets.
\end{abstract}

\section{INTRODUCTION}
Given element sets $X_1,\ldots,X_n$ ($n\ge 2$), the problem of finding consistent pairwise bijections between all pairs of sets is known as multi-way matching. As a critical problem in computer science, it is widely used in many computer vision tasks, such as object recognition \citep{demirci2006object, yang2011multiple}, shape analysis \citep{petterson2009exponential}, and structure from motion \citep{dai2014simple, roberts2011structure}. It can also be applied to other fields (e.g. multiple graph matching \citep{lacoste2006word, yan2013joint} and data source integration \citep{zhang2015principled}).

In most cases, the multi-way matching problem is approached as a weighted multi-dimensional matching optimzation or relaxation (e.g. the works \citep{pachauri2013solving, yan2015matrix, yan2015consistency}). This objective is easy to solve when $n=2$, since no consistency between matchings is required. However, as $n\ge 3$, the problem becomes hard due to the combinatorial constraints induced by consistency and we can show that the underlying optimization is NP-hard to solve in general. Therefore, approximate methods, such as convex relaxation \citep{pachauri2013solving} and matrix decomposition \citep{yan2015matrix} have been proposed. These methods work well on datasets with little noise but may become unreliable when more realistic noise levels are present in the data.

In this paper, we aim to find algorithms that directly optimize the true objective of the weighted multi-dimensional matching problem. One intuitive approach is to iteratively update the matching between pairs of sets $(X_i,X_{i+1})$'s for $i=1,\ldots,n-1$, but this may produce significant errors once one erroneous pairwise matching is found in the iterative process\citep{le2007bundle, tsochantaridis2005large, volkovs2012efficient}. Alternatively, one can simply perform coordinate updates on the objective since each coordinate update subproblem is a weighted bipartite matching which can be efficiently solved optimally. However, coordinate update approaches depends heavily on good initialization and may produce bad performance due to local optima.

In this paper, we combine the above ideas and design an effective method for the multi-way matching problem. We build an undirected graph with edge weights from all pairwise matching similarity values, and use its Maximum Spanning Tree (MST) to find a good order for computing $n-1$ pairwise matchings. This helps avoid bad local optima since it focuses initially on more reliable matchings in the coordinate updates. This seemingly simple idea yields good performance in practice while also enjoying theoretical guarantees. Similar ideas have been discussed in previous works (e.g. \citep{yan2016multi}), but lacked a comprehensive theoretical analysis. In real experiments, we obtain surprisingly strong results on many well-known datasets. For instance, we reliably get $\bm{0\%}$ error on the famous datasets \textbf{CMU House} and \textbf{CMU Hotel} for the task of stereo landmark alignments with $m=30$ points (which has not been easy for previous algorithms to achieve). Theoretically, we not only guarantee that our algorithm solves the problem optimally when pairwise alignment methods work but we {\em also} guarantee optimality with high probability when a spanning tree on the noise parameter graph has small bottleneck weight (the largest weight in a spanning tree) after imposing some other mild assumptions.

\section{CONSISTENT MATCHING FOR SETS OF ELEMENTS}
\label{sec:preliminary}
We frame the multiway matching problem as described by \citet{pachauri2013solving}. Assume we have $n$ element sets $X_1,X_2,\ldots,X_n$ where each $X_i$ contains $m$ elements $X_i=\{x^i_1,x^i_2,\ldots,x^i_m\}$. For any pair of element sets $(X_i, X_j)$, we assume that there exists a bijection between their elements such that element $x^i_p$ is mapped to element $x^j_q$ if they are similar to each other. For example, $X_1,\ldots,X_n$ could be $n$ images of an everyday object (say a chair) and each image $X_i$ therein contains $m$ pixels $x^i_1,x^i_2,\ldots,x^i_m$. Since these images describe the same object type, we expect a bijection to exist between the parts (or pixels) within the pairs of images.

Clearly, such bijections should be consistent with each other. In other words, if element $x^i_p$ is mapped to element $x^j_q$ and element $x^j_q$ is mapped to element $x^k_r$, then element $x^i_p$ should be mapped to element $x^k_r$. More specifically, given the element sets $X_1$, .., $X_n$, we are interested in finding a consistent bijection $\tau_{ij}:\{1,\ldots,m\}\rightarrow\{1,\ldots,m\}$ between each pair of element sets $(X_i,X_j)$ such that: $x^{j}_{\tau_{ij}(p)}$ is mapped to $x^i_p$, $\tau_{ii}$ is the identity transform, $\tau_{ij}=\tau_{ji}^{-1}$ and $\tau_{jk}\circ\tau_{ij}=\tau_{ik}$, for any element sets $X_i$, $X_j$, $X_k$ and any element $x^i_p$.

Achieving the above is equivalent to reordering the elements in each element set $X_i$ such that the elements with the same index correspond to each other. Mathematically, finding a consistent bijection $\tau_{ij}$ for each pair of element sets $(X_i,X_j)$ is equivalent to finding a bijection $\sigma_i:\{1,\ldots,m\}\rightarrow\{1,\ldots,m\}$ for each element set $X_i$, such that element $x^i_p$ is mapped to element $x^j_q$ if and only if $\sigma_i(p)=\sigma_j(q)$. We easily see that these mappings satisfy $\tau_{ij}=\sigma_j^{-1}\circ\sigma_i$ for any $\tau_{ij}$, $\sigma_i$ and $\sigma_j$.

In order to find the mappings $\sigma_i$'s, \citet{pachauri2013solving} proposed an alternative objective function. They assume that we are given a \textit{similarity matrix} $T_{ij}\in\mathbb R^{m\times m}$ for each pair of sets $(X_i,X_j)$. The entry $[T_{ij}]_{p,q}$ in the $p^{th}$ row and $q^{th}$ column of $T_{ij}$ represents the similarity level between elements $x^i_p$ and $x^j_q$. The closer two elements are each other, the larger this similarity level is. By symmetry, we also require that $T_{ij}=T_{ji}^\top$ for any pair of $(T_{ij},T_{ji})$. Without loss of generality, we will assume that $[T_{ij}]_{p,q}$ are constrained to the range $[0,1]$. Ideally, elements $x^i_p$ and elements $x^j_q$ can be perfectly matched to each other if $[T_{ij}]_{p,q}=1$ and is maximal. We also hope to avoid matching pairs of elements that not related to each other,  e.g. when $[T_{ij}]_{p,q}=0$ or is minimal. \citet{pachauri2013solving} recovered the mappings $\sigma_1,\ldots,\sigma_n$ by solving the following optimization problem:
\begin{equation}
\max\limits_{\sigma_1,\ldots,\sigma_n}\mathcal L(\sigma_1,\ldots, \sigma_n):=\sum\limits_{i=1}^n\sum\limits_{j=1}^n\langle P(\sigma_j^{-1}\circ\sigma_i), T_{ij}\rangle
\label{obj:v1}
\end{equation}
where $P(\sigma)\in\mathbb R^{m\times m}$ is a permutation matrix satisfying 
$$[P]_{p,q}=
\begin{cases}
1& \text{if }\sigma(p)=q\\
0& \text{otherwise.}
\end{cases}$$

Notice that all permutation matrices are orthogonal matrices and $P(\sigma_j^{-1}\circ\sigma_i)=P(\sigma_i)^{-1}P(\sigma_j)$ for any mappings $\sigma_i$ and $\sigma_j$. Denote the set of all $m\times m$ permutation matrices as $\mathcal P_m$. We rewrite the objective function in Equation \eqref{obj:v1} as 
\begin{equation}
\max\limits_{A_1,\ldots,A_n\in\mathcal P_m}\mathcal L(A_1,\ldots, A_n):=\sum\limits_{i=1}^n\sum\limits_{j=1}^n \text{tr}(A_iT_{ij}A_j^\top)
\label{obj:v2}
\end{equation}
since $A_i^\top= A_i^{-1}$, where $A_i=P(\sigma_i)$ is a permutation matrix for the element set $X_i$. Note that the solution for this optimization problem is not unique (in fact, it has at least $m!$ different tuples of solutions) since $(A_1,\ldots,A_n)=(P\hat A_1,\ldots,P\hat A_n)$ is an optimal solution if $(A_1,\ldots,A_n)=(\hat A_1,\ldots,\hat A_n)$ is, for any permutation matrix $P\in\mathcal P_m$.

A naive method for solving this problem is to recover $A_1,\ldots,A_n$ from equations $P_{ij}=A_i^\top A_j$, where $P_{ij} = \argmax\limits_{P\in\mathcal P_m}\text{tr}(P^\top T_{ij})$, for all $i,j\in\{1,\ldots,n\}$. We call this method \textit{Pairwise Alignment}. It clearly does not always work since the matrices $P_{ij}$'s may not be consistent with each other (i.e. they don't always satisfy $P_{ij}P_{jk}=P_{ik}$) and may not correspond to a solution for $(A_1,\ldots,A_n)$. In the next section, we will propose novel algorithms that solve this optimization problem.

\section{COORDINATE OPTIMIZATION WITH SMART INITIALIZATION}
\label{sec:original}
The optimization problem in Equation \eqref{obj:v1} is essentially a maximum weighted $n$-way matching problem, which is NP-hard to solve (the proof is in \S \ref{sec:proof-np-hard} of the supplementary material). Therefore, we cannot find solutions to this problem for arbitrary input values of $T_{ij}$. Instead, we will constrain the similarity matrices that are used as inputs to the problem. To approximate the problem, \citet{pachauri2013solving} proposed an eigenvalue decomposition-based method by first relaxing the combinatorial optimization into a continuous one and then rounding the solution using the Kuhn-Munkres algorithm \citep{kuhn2010hungarian}. However, \citet{pachauri2013solving} could only guarantee their solution when every similarity matrix $T_{ij}$ was close to the ground truth permutation matrix $P(\tilde\sigma_j^{-1}\circ\tilde\sigma_i)$ (denote $\tilde\sigma_1,\ldots,\tilde\sigma_n$ to be the ground truth mappings we want to find). Unfortunately, this is rarely the case in practice. In the next section, we will present a more general method for solving this problem via coordinate ascent.

\subsection{Coordinate ascent over permutations}
Consider the objective function in Equation \eqref{obj:v2}. For each permutation matrix $A_i$, since $\text{tr}(A_iT_{ii}A_i^\top)=\text{tr}(A_i^\top A_iT_{ii})=\text{tr}(T_{ii})$ is a constant, and $\text{tr}(A_iT_{ij}A_j^\top)=\text{tr}(A_jT_{ij}^\top A_i^\top)=\text{tr}(A_jT_{ji} A_i^\top)$ for any permutation matrix $A_j$, we know that, if we fix all of the other permutation matrices $A_1,\ldots,A_{i-1},A_{i+1},\ldots,A_n$, then the maximization problem becomes
\begin{equation}
\begin{aligned}
&\argmax\limits_{A_i\in\mathcal P_m}\mathcal L(A_1,\ldots,A_n)=\argmax\limits_{A_i\in\mathcal P_m}\text{tr}(A_i^\top \sum\limits_{1\le j\le n, i\ne j}A_jT_{ij}).
\label{eqn:coor-update}
\end{aligned}
\end{equation}

The optimization in Equation \eqref{eqn:coor-update} can be solved in polynomial time (for example, through the $O(m^3)$ Kuhn-Munkres algorithm \citep{kuhn2010hungarian}). Hence, a naive coordinate algorithm is easy to derive: initialize the permutation matrices $A_1,\ldots,A_n$ (either randomly or deterministically). Then, for each iteration, randomly pick $i\in\{1,\ldots,n\}$, update $A_i$ according to Equation \eqref{eqn:coor-update}, and repeat until convergence. Unfortunately, standard ways of initializing such an algorithm lead to poor local optima (see \S \ref{sec:expr-CMU}). Better performance can be achieved, however, if we use pairwise alignment information to construct a good initialization. This approach is discussed in \S \ref{sec:MST_initialization}.

\subsection{MST-based initialization}
\label{sec:MST_initialization}
We seek a good initialization for the coordinate update approach summarized in Equation \eqref{obj:v1}. Consider a single term $\text{tr}(A_iT_{ij}A_j^\top)$ in the objective function in Equation \eqref{obj:v2}. As we maximize that objective, we say that we are {\em confident} in the values chosen for $A_i$ and $A_j$ if the corresponding term  $\text{tr}(A_iT_{ij}A_j^\top)$ is large. Define  
\begin{equation}
f(T_{ij}):=\max\limits_{P\in\mathcal P_m}\text{tr}(P^\top T_{ij})=\max\limits_{A_i,A_j\in\mathcal P_m}\text{tr}(A_iT_{ij}A_j^\top)
\end{equation}
for each $T_{ij}$. We call an initialization of our algorithm {\em convincing} if $A_i=\hat A_i$ and $A_j=\hat A_j$ and $\text{tr}(\hat A_iT_{ij}\hat A_j^\top)$ is close or equal to $f(T_{ij})$ for some permutation matrices $\hat A_i,\hat A_j\in\mathcal P_m$.

The above intuition encourages us to first initialize the matrices $A_i$, $A_j$ that correspond to values of $f(T_{ij})$ that are large. To achieve this, we build an undirected graph $G=(V,E)$ where each element set $X_i$ corresponds to one vertex $v_i\in V$ and each pair of element sets $(X_i, X_j)$ ($i\ne j$) corresponds to an edge $(v_i, v_j)\in E$ with weight $f(T_{ij})=f(T_{ji})$. We then find a Maximum Spanning Tree $T=(V, E')$ of $G$. Then, we initialize the matrices $A_1,\ldots,A_n$ along the edges in $E'$ as follows. Initially, we have $n$ sets $S_1,\ldots,S_n$ of vertices, each containing one vertex in $V$. Then, for every edge in each edge $(v_i, v_j)\in E'$, we try to {\em combine} them together and use the similarity $T_{ij}$ to find the permutation matrices corresponding to vertices in the sets contain $v_i$ and $v_j$. Details are shown in Algorithm \ref{alg:original}. 

\begin{algorithm}[htb]
\caption{MST-based coordinate updates for multi-way matching}
\label{alg:original}
\begin{algorithmic}[1]
\REQUIRE ~~\\
    The similarity matrices $T_{ij}$'s ($i,j\in\{1,\ldots,n\}$).

\ENSURE ~~\\

    The permutation matrices $A_i$'s ($i\in\{1,\ldots,n\}$).

\STATE Construct graph $G=(V,E)$ as in \S \ref{sec:MST_initialization}, compute a Maximum Spanning Tree $T=(V, E')$ of $G$;
\STATE Initialize $S_i\leftarrow\{v_i\}$ for each $v_i\in V$; For each $A_i$, initialize it to be any permutation matrix in $\mathcal P_m$;
\FOR{each edge $(v_i, v_j)\in E'$}
\STATE Compute $\hat P=\argmax\limits_P \text{tr}(P^\top A_iT_{ij}A_j^\top)$;
\STATE Update $A_{j'}\leftarrow \hat PA_{j'}$ for each $v_{j'}\in S_j$;
\STATE Let $S'=S_i\cup S_j$;
\STATE Update $S_k\leftarrow S'$ for each $v_k\in S'$;
\ENDFOR
\WHILE {Not converged}
\STATE Randomly pick $i\in\{1,\ldots,n\}$;
\STATE Update $A_i$ according to Equation \eqref{eqn:coor-update};
\ENDWHILE

\end{algorithmic} 
\end{algorithm} 

The above algorithm uses a Maximum Spanning Tree to initialize the permutation matrices $A_1,\ldots,A_n$. To iteratively {\em combine} the vertices in $V$ to find an initialization, we need each edge $(v_i, v_j)$ that is selected to have a relatively large $f(T_{ij})$ value. The Maximum Spanning Tree of $G$ achieves this. Subsequently, the algorithm above simply iterates the usual coordinate update process. We will next analyze how this initialization provides a reliable starting point for the coordinate updates that will ultimately produce a good final set of permutation matrices $(A_1,\ldots,A_n)$.

\subsection{Analysis of the coordinate update}
\label{sec:analysis}
\subsubsection{Analysis without noise}
\label{sec:anaysis-no-noise}
We now analyze the behavior of Algorithm \ref{alg:original}. First, consider a simple case where we guarantee through the Pairwise Alignment method that there are {\em consistent} permutation matrices $P_{ij}=f(T_{ij})$ for each pair of element sets $(X_i, X_j)$, i.e. $P_{ij}P_{jk}=P_{ik}$ for all $i,j,k\in\{1,\ldots,n\}$ (here by guarantee we mean that the maximum value of $\text{tr}(P^\top T_{ij})$ will be achieved for some unique permutation matrix $P$, for each similarity matrix $T_{ij}$). If we have consistency, then we can easily recover the optimal $(A_1,\ldots,A_n)$ from the $P_{ij}$ matrices by setting $A_i=P_{1i}$ for each $A_i$ since each single term $\text{tr}(A_iT_{ij}A_j^\top)=\text{tr}(A_j^\top A_iT_{ij})=\text{tr}(P_{ij}^\top T_{ij})$ in the objective function in the Equation \eqref{obj:v2} is maximized.

What is Algorithm \ref{alg:original}'s behavior under this constraint? Can we guarantee that it recovers all $A_i$'s matrices optimally? The answer is YES. We leverage the following theorem:
\begin{theorem}
\label{thm:no-noise}
If we recover consistent permutation matrices $P_{ij}=f(T_{ij})$ for all pairs of element sets $(X_i,X_j)$ using the Pairwise Alignment method, then we can guarantee that Algorithm \ref{alg:original} solves the optimization problem in Equation \ref{obj:v2} optimally. 
\end{theorem}
The proof is in \S \ref{sec:proof1} of the supplementary material. From Theorem \ref{thm:no-noise}, we know that Algorithm \ref{alg:original} is at least as good as the Pairwise Alignment method. Moreover, the optimality cases in Theorem \ref{thm:no-noise} subsume all cases that \citet{pachauri2013solving} claimed they could solve optimally. Next, we go even further and guarantee optimality in much more general settings.

\subsubsection{Analysis with noise}

A more interesting setting is when the matrices $T_{ij}$'s are not perfect and consistent permutation matrices but rather have been corrupted by noise. If we denote the optimal solution for the optimization problem in Equation \eqref{obj:v2} as $(A_1,\ldots,A_n)=(\hat A_1,\ldots,\hat A_n)$, then ideally the best input data we could have for each $T_{ij}$ would be $T_{ij}=\hat T_{ij}:=\hat A_i^\top\hat A_j$. In the case where $T_{ij}=\hat T_{ij}$ for each $i,j\in\{1,\ldots,n\}$, it is obvious from Theorem \ref{thm:no-noise} that Algorithm \ref{alg:original} solves this optimization problem optimally.

What if the similarity matrices have noise and $T_{ij}$ is not perfectly equal to $\hat T_{ij}$ for some (or all) of the $T_{ij}$'s? To analyze Algorithm \ref{alg:original}, we will assume that the $T_{ij}$ inputs are random perturbations near $\hat T_{ij}$. We only need to consider matrices $T_{ij}$ where $i\ne j$ since the Algorithm does not depend on $T_{ii}$ in any way. Recall that we assumed that the entries of $T_{ij}$ ranged from $[0, 1]$. We propose the following model of the noise that generates the entries of $T_{ij}$ as perturbations of the ground-truth $\hat T_{ij}$:
\begin{equation}
[T_{ij}]_{p,q}=
\begin{cases}
1-Z^2_{ijpq}& \text{if }i< j\text{ and }[\hat T_{ij}]_{p,q}=1\\
Z^2_{ijpq}& \text{if }i< j\text{ and }[\hat T_{ij}]_{p,q}=0\\
[T_{ji}]_{q,p}& \text{if }i>j.
\end{cases}
\label{eqn:model}
\end{equation}
Here $Z_{ijpq}\sim \mathcal N(0, \eta_{ij})$'s are independent Gaussian random variables for any $1\le i< j\le n$ and $p,q\in\{1,\ldots,m\}$. We assume that different $T_{ij}$ matrices may have different variance parameters $\eta_{ij}$'s since we may have different noise levels for different pairs of element sets. Also, we require $\eta_{ij}\le O(1)$ for each $\eta_{ij}$ since we want the similarity matrices to only have entries in $[0,1]$. Notice that we still maintain $T_{ij}=T_{ji}^\top$ for all $i\ne j$ under the model in Equation \eqref{eqn:model}. We now have the following more general theorem:

\begin{theorem}
\label{thm:noise}
With probability $1-o(1)$ and for sufficiently large $n$ and $m$,
Algorithm \ref{alg:original} finds an optimal solution for the optimization problem in Equation \eqref{obj:v2} under the following conditions:
\begin{itemize}
\item $n\ge20\ln m$, and $\exists\gamma>0$ such that $n\le m^\gamma$,
\item the bottleneck length of the \texttt{Minimum Bottleneck Spanning
    Tree} of $G$ is at most $\frac{1}{4(3+\gamma)\ln m+4}$ where
  $G=(V, E)$ is a complete undirected weighted graph, with a vertex $v_i\in V$ for each set $X_i$ and with edges $(v_i,v_j)\in E$ with weight $\eta_{ij}$,
\item and $\max\limits_{1\le i < j\le n}\eta_{ij}\le\frac13$.
\end{itemize}
\end{theorem}

The proof is in \S \ref{sec:proof2} of the supplementary material. Here the Minimum Bottleneck Spanning Tree of a graph $G$ means a spanning tree of $G$ which has minimal edge weight on its heaviest edge. It seems that our algorithm could work well if $n$ and $m$ are both large and there exists a spanning tree of graph $G$ with all edge weights no more than $O(\frac{1}{\log m})$. In the proof for this theorem we will show that we can use the Pairwise Alignment method to solve the optimization problem optimally with high probability if all edges of $G$ have weight no more than $O(\frac{1}{\log m})$. This is the same guarantee asymptotically as our bottleneck weight bound but the latter applies for all edges in $E$. So, our algorithm remains optimal (with high probability) for a much broader set of inputs.

\subsection{Practical improvements}
\label{sec:improvements}

In \S \ref{sec:MST_initialization} and \ref{sec:analysis}, we introduced our algorithm and discussed its theoretical guarantees. However, to make Algorithm \ref{alg:original} better in practice, we also suggest some minor improvements that tend to provide slightly better empirical performance.

\subsubsection{Combining initialization with coordinate optimization}
In Algorithm \ref{alg:original}, we propose a coordinate update process after an initialization step. However, there is a possibility that we may find bad solutions under this initialization as well. Therefore, it is helpful to add a coordinate update process right after each iteration of initialization that may potentially fix some errors the algorithm made during that iteration. During each iteration, after we have processed the vertices in the set $S'$ (line 6 of Algorithm \ref{alg:original}), we can do a coordinate update on the corresponding permutation matrices $A_k$'s where $v_k\in S'$ as:
\begin{equation}
A_k=\argmax\limits_{P}\text{tr}(P^\top\sum\limits_{v_{k'}\in S',k'\ne k}A_{k'}T_{kk'}).
\label{eqn:coor-update2}
\end{equation}
By adding these intermediate update steps, we no longer need to have a final coordinate update step since the additional coordinate updates after the last iteration of initialization have already played that role.

\subsubsection{Using a good MST edge ordering}
In Algorithm \ref{alg:original}, we performed initialization by enumerating the edges of the Maximum Spanning Tree $T$. It is reasonable that running the updates in a good order along the edges may be beneficial. In this section, we propose two kinds of ordering that we have found work well in practice: Prim's order and Kruskal's order.

As in line 5 of Algorithm \ref{alg:original}, we need to update $|S_j|$ different permutation matrices in one step. Even though we have proved that this algorithm works well in many cases, it can be improved if we are more {\em cautious} and update fewer permutation matrices at each iteration. On way is to use Prim's algorithm \citep{ahuja1988network} to compute the Maximum Spanning Tree and then process the edges in the order that we get them through the execution of Prim's algorithm. Since there is only one vertex in the set $|S_j|$ each time, we only need to update one permutation matrix at each iteration. We call this ordering \textit{Prim's order}.

Alternatively, the edge weights themselves are potentially important for initialization. As discussed in \S \ref{sec:anaysis-no-noise}, we are more confident in edges $(v_i, v_j)$ whose weights $f(T_{ij})$'s are large. Therefore, we update according to edges that we trust more first. To achieve that goal, we can process the edges in the descending weight order. This is exactly the edge order that we get from running Kruskal's algorithm \citep{ahuja1988network} . We call this ordering \textit{Kruskal's order}.

\subsubsection{The overall algorithm}
\begin{algorithm}[htb]
\caption{Improved MST-based coordinate updates for multi-way matching}
\label{alg:improved}
\begin{algorithmic}[1] 
\REQUIRE ~~\\
    The similarity matrices $T_{ij}$'s ($i,j\in\{1,\ldots,n\}$).

\ENSURE ~~\\

    The permutation matrices $A_i$'s ($i\in\{1,\ldots,n\}$).

\STATE Construct the Graph $G=(V,E)$ as in the \S \ref{sec:MST_initialization}, Compute a Maximum Spanning Tree $T=(V, E')$ of $G$;
\STATE Sort the edges in $E'$ with Prim's order or Kruskal's order as discussed in \S \ref{sec:improvements};
\STATE Initialize $S_i\leftarrow\{v_i\}$ for each $v_i\in V$; For each $A_i$, initialize it to be any permutation matrix in $\mathcal P_m$;
\FOR{each edge $(v_i, v_j)\in E'$}
\STATE Compute $\hat P=\argmax\limits_P \text{tr}(P^\top A_iT_{ij}A_j^\top)$;
\STATE Update $A_{j'}\leftarrow \hat PA_{j'}$ for each $v_{j'}\in S_j$;
\STATE Let $S'=S_i\cup S_j$;
\STATE Update $S_k\leftarrow S'$ for each $v_k\in S'$;
\WHILE {Not converged}
\STATE Randomly pick $v_k\in S'$;
\STATE Update $A_k$ according to Equation \eqref{eqn:coor-update2};
\ENDWHILE
\ENDFOR
\end{algorithmic} 
\end{algorithm} 

By adding the heuristics mentioned above, we obtain a slight modification of our algorithm as shown in Algorithm Box \ref{alg:improved}. This algorithm works slightly better and we will explore how these heuristics fare in the experiments section. Using techniques similar to those in the proof of Theorem \ref{thm:no-noise}, it is easy to show that Algorithm \ref{alg:improved} is at least as good as the Pairwise Alignment method:
\begin{theorem}
\label{thm:no-noise2}
If we can guarantee the recovery of pairwise-consistent permutation matrices $P_{ij}=f(T_{ij})$ for each pair of element sets $(X_i,X_j)$ using the Pairwise Alignment method, then we can guarantee that Algorithm \ref{alg:improved} solves the optimization problem in Equation \ref{obj:v2} optimally. 
\end{theorem}

\section{EXPERIMENTS}

In this section, we will show how our algorithms behave in practice. We focus primarily on computer vision datasets. For each dataset, we compare our algorithm with the Permutation Synchronization algorithm \citet{pachauri2013solving}, which is a state-of-the-art method for multi-way matching.
\subsection{PCA reconstruction of MNIST digits}
\label{sec:experiments}
\begin{figure*}[ht]
\centering
\vspace{.417in}
 \subfigure[The PCA reconstruction errors of our methods compared to two baseline methods. The horizontal axis represents the reduction dimensionality $k$. The performance of the two versions of Algorithm \ref{alg:improved} are almost the same, and both clealy outperform previous approaches. ]{
    \label{fig:err} 
    \includegraphics[trim = 15mm 60mm 10mm 94mm, width=0.47\textwidth]{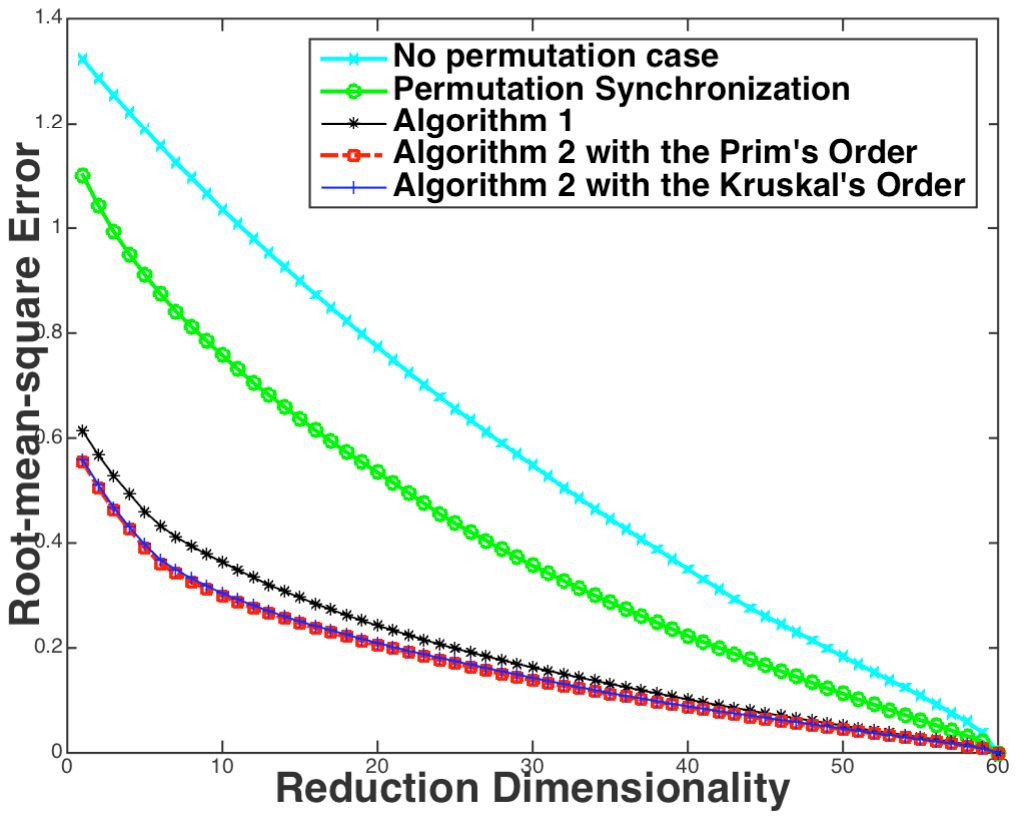}}
    \hspace{.1in}
  \subfigure[The reconstructed results of an image of digit 0 by all methods. For each figure, we show the reconstructed images with $k=4,11,18,\ldots, 60$ eigenvectors, respectively. We see that our methods can reconstruct the image well even at $k=4$. Meanwhile, the other two methods require many more eigenvectors to reconstruct a recognizable digit 0 image. ] {
    \label{fig:subfig:b} 
    \includegraphics[trim = 8mm 57mm 21mm 200mm, width=0.47\textwidth]{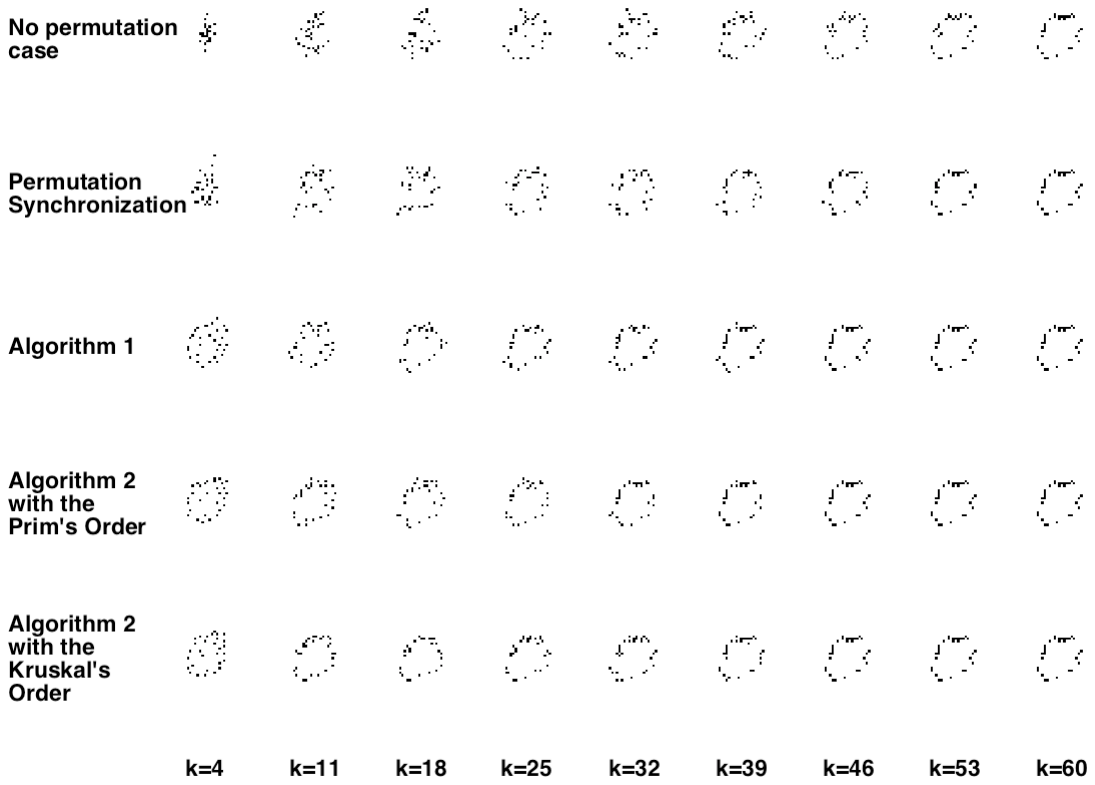}}
  \label{fig:digit} 
\vspace{-.078in}
\caption{Results for the PCA reconstruction experiment}
\label{fig:PCA}
\end{figure*}
Our first experiment is on image compression and recovery. We use the \textbf{MNIST} dataset \citep{lecun1998gradient}, which contains 70,000 images of individual handwritten digits from $\{0,\ldots, 9\}$.

In one experiment, we randomly selected $n=100$ images $I_1,\ldots,I_n$ from the dataset, where each digit has roughly $\frac n{10}$ images. Note that we do not use a larger number of images because of scaleability limitations of \citet{pachauri2013solving} which we need as our baseline in the evaluation. Our algorithms, however, easily scale to much larger data-sets. The goal of this experiment is to compress the MNIST images with low dimensionality. We represent each image $I_i$ as an element set by randomly selecting $m=30$ white pixels (the MNIST digits are white drawings on a black background). This forms the element set $X_i=\{x^i_{j}=(a^i_j,b^i_j):j\in\{1,\ldots,m\}\}$, where $(a^i_j,b^i_j)$ is the coordinate of the $j^{th}$ selected pixel of image $I_i$.

We will use Principal Components Analysis (PCA) as our compression technique and apply it to the element sets $X_1,\ldots,X_n$. We can view each $X_i$ as a matrix $Y_i\in\mathbb R^{m\times 2}$ where the $j^{th}$ row of $Y_i$ is merely $(a^i_j, b^i_j)$. It is straightforward to vectorize these $Y_i$ matrices and apply PCA to compressively store the vectors $\text{vec}(Y_1),\ldots,\text{vec}(Y_n)\in\mathbb R^{2m}$. We explore various levels of compression by keeping only the $k\in\{1,\ldots,\min(n,2m)\}$ leading eigenvectors from PCA as well as a mean vector in $\mathbb R^{2m}$. However, since $X_1,\ldots,X_n$ are element sets, we are free to permute the order of the element sets prior to the application of PCA. Since each of the MNIST images is a digit, it is reasonable to believe that for many pairs of images, there should be a bijection or correspondence relationship between the pixels that compose each digit. Intuitively, permutations that discover that bijection will help improve PCA performance. We will evaluate how various permutation algorithms that precede PCA help its ability to reconstruct the original data at various compression levels $k$.  Our goal is to reduce the reconstruction error by using our algorithms to reorder the elements in each element set $X_i$ prior to PCA compression.

Results are shown in Figure \ref{fig:PCA}. Here we use the radial basis function (RBF) kernel to compute the similarity matrices $T_{ij}$'s. More specifically, $[T_{ij}]_{p,q}=\exp(-\frac{||x^i_p-x^j_q||^2}{2\sigma^2})$ where $\sigma$ is a parameter. This function is suitable for our settings since we expect $[T_{ij}]_{p,q}$ to be close to 1 when $x^i_p$ and $x^j_q$ are close to each other and to be close to 0 otherwise. We compare the two versions of our algorithms with two baseline cases. In the first baseline, we do not reorder the pixels (the no permutation case). In the second baseline, we use the Permutation Synchronization algorithm to reorder the pixels before compression. For all methods in our experiment, we selected the best $\sigma$ value that achieves the best performance. Figure \ref{fig:PCA} shows the reconstruction errors and the recovered digits obtained from all five methods. All versions of our algorithm outperform the two baseline methods. Note that both versions of Algorithm \ref{alg:improved} slightly outperform Algorithm \ref{alg:original} (although they often produce similar results) which is due to the additional heuristics we discussed in \S \ref{sec:improvements}.

\subsection{Stereo landmark alignment}
\label{sec:expr-CMU}
\begin{table*}[ht]
\caption{Average Error Rates of alignments for the datasets House, Hotel, Building and Sentence} \label{tab:aer}
\begin{center}
\begin{tabular}{llll}
{\bf TASK}  &{\bf PERM-SYNC} &{\bf PRIM'S ORDER} &{\bf KRUSKAL'S ORDER} \\
\hline \\
CMU House RBF &$(22.61\pm7.49)\%$ &$\bm{(0.00\pm0.00)\%}$ &$\bm{(0.00\pm0.00)\%}$\\
CMU House Alignment &$(4.71\pm1.98)\%$& $\bm{(0.81\pm0.96)\%}$& $(1.89\pm2.21)\%$\\
CMU Hotel RBF & $(18.63\pm2.90)\%$& $\bm{(0.00\pm0.00)\%}$&$\bm{(0.00\pm0.00)\%}$ \\
CMU Hotel Alignment &$(5.22\pm2.55)\%$\ & $\bm{(3.59\pm0.75)\%}$& $(4.20\pm0.71)\%$\\
Building RBF & $(86.71\pm3.36)\%$ & $\bm{(49.87\pm0.24)\%}$& $(50.39\pm0.25)\%$\\
Building Alignment & $(50.49\pm1.09)\%$ & $\bm{(48.52\pm0.50)\%}$& $(48.61\pm0.52)\%$\\
Sentence RBF & $(62.65\pm2.90)\%$ & $\bm{(55.26\pm0.82)\%}$& $(55.85\pm0.95)\%$\\
Sentence Alignment & $(58.69\pm2.99)\%$ & $(56.06\pm1.19)\%$& $\bm{(55.59\pm1.58)}\%$
\end{tabular}
\end{center}
\end{table*}
The second computer vision task we focus on is stereo matching. The goal is to align pixels from multiple images of a single object where the images are taken from various vantage points. For this experiment, we have 2 datasets, the \textbf{CMU House} dataset and the \textbf{CMU Hotel} dataset\footnote{\href{http://vasc.ri.cmu.edu/idb/html/motion/}{http://vasc.ri.cmu.edu/idb/html/motion/}.}. Each of them contains $n$ images of the same toy house ($n=111$ for the CMU House dataset and $n=101$ for the CMU Hotel dataset). For each of the two toy houses, we have $m=30$ landmark points selected, and each of the $n$ figures contains a different view of these $m$ landmark points. Our goal is to find a consistent mapping that aligns points that correspond to the same landmarks together.

The element sets are constructed by extracting visual features. Once again, we use the RBF kernel to compute similarity matrices. We denote the tasks in this experiment as \textit{CMU House RBF} and \textit{CMU Hotel RBF}.  Since we have the ground truth alignments for these two datasets, it is also reasonable to use some local alignments between pairs of figures to construct the similarity matrices. Hence, we also use the outputs of the Pairwise Alignment method as our similarity matrices. We call these corresponding tasks \textit{CMU House Alignment} and \textit{CMU Hotel Alignment}. To evaluate performance, we compute the average error rates of all pairs of element sets.

Results are shown in the first four lines of Table \ref{tab:aer}. For this experiment, we only report performance from Algorithm \ref{alg:improved} though our other algorithm performs almost as well. We compare the two versions of our proposed algorithm with the Permutation Synchronization method. Here we show results over 10 trials of the experiments since performance is stochastic due to the random re-ordering of the images and the pixels prior to input to the algorithms. We can see that both our methods reliably solve this problem with $\textbf{100\%}$ accuracy in the RBF setting. Meanwhile the baseline method behaves much worse. For the Alignment setting, the baseline method's behaves better since it tends to excel with those types of inputs. Nevertheless, our methods still outperform it, especially the Prim's Order version.

Notice that the initialization component of our algorithms is very important. Without the smart initialization technique, coordinate updates behave unreliably. For instance, in the CMU House RBF task, we obtain an average error rate of $(3.90\pm2.63)\%$ if we randomly initialize all permutation matrices. Meanwhile we always get $0\%$ error rate when we use our MST-based initialization technique.

\subsection{Repetitive structures of key points}
\begin{figure*}[ht]
\centering
\vspace{-.245in}
 \subfigure[The Permutation Synchronization method under the RBF setting]{
    \label{fig:rbf-sync} 
    \includegraphics[trim = 8mm 95mm 8mm 83mm, width=0.316\textwidth]{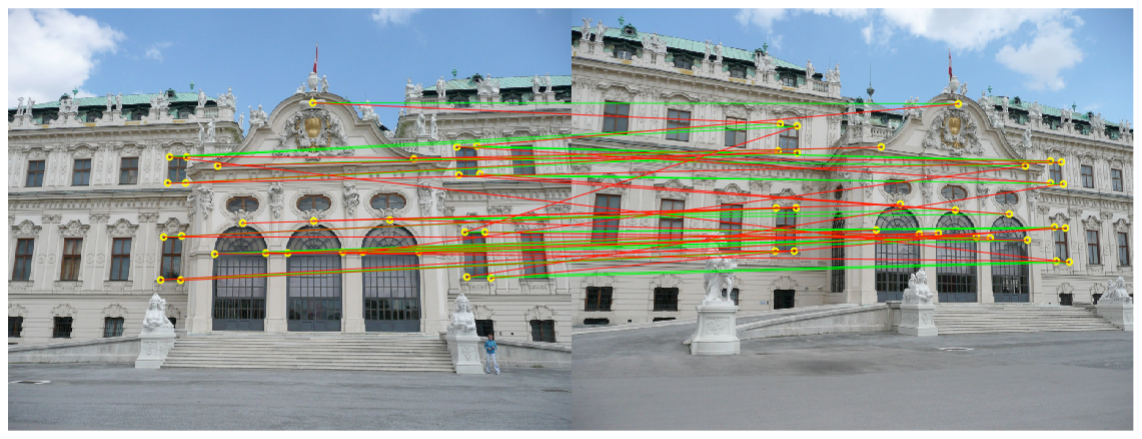}}
    \hspace{.00in}
  \subfigure[Algorithm \ref{alg:improved} with Prim's Order under the RBF setting]{
    \label{fig:subfig:b} 
    \includegraphics[trim = 8mm 95mm 8mm 83mm, width=0.316\textwidth]{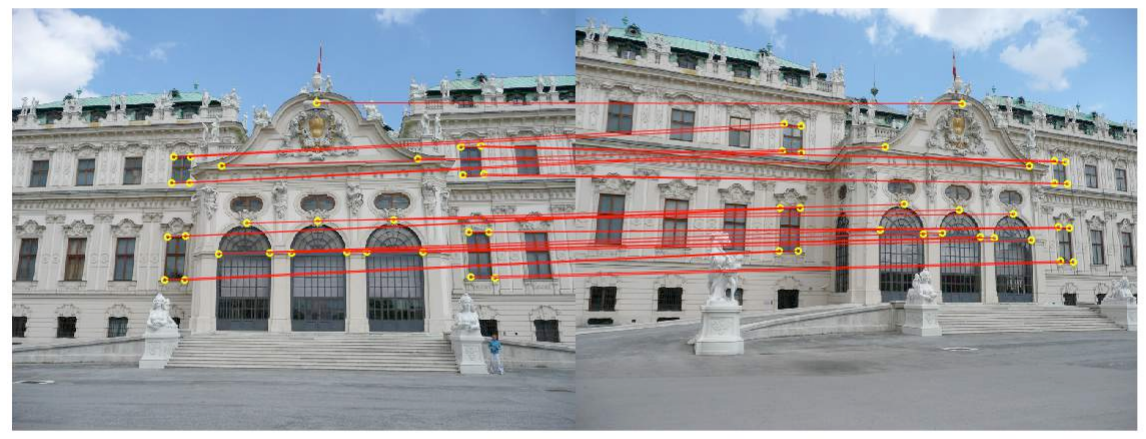}}
  \label{fig:rbf-prim} 
  \hspace{.00in}
  \subfigure[Algorithm \ref{alg:improved} with Kruskal's Order under the RBF setting]{
    \label{fig:subfig:b} 
    \includegraphics[trim = 8mm 95mm 8mm 83mm, width=0.316\textwidth]{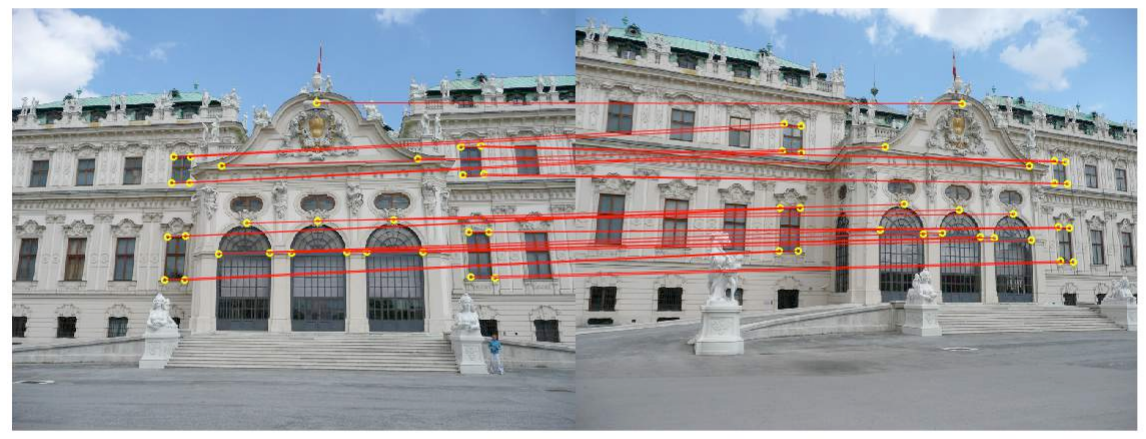}}
  \label{fig:rbf-kruskal} 
  \vspace{-.127in}
  \subfigure[The Permutation Synchronization method under the Alignment setting]{
    \label{fig:rbf-sync} 
    \includegraphics[trim = 8mm 95mm 8mm 83mm, width=0.316\textwidth]{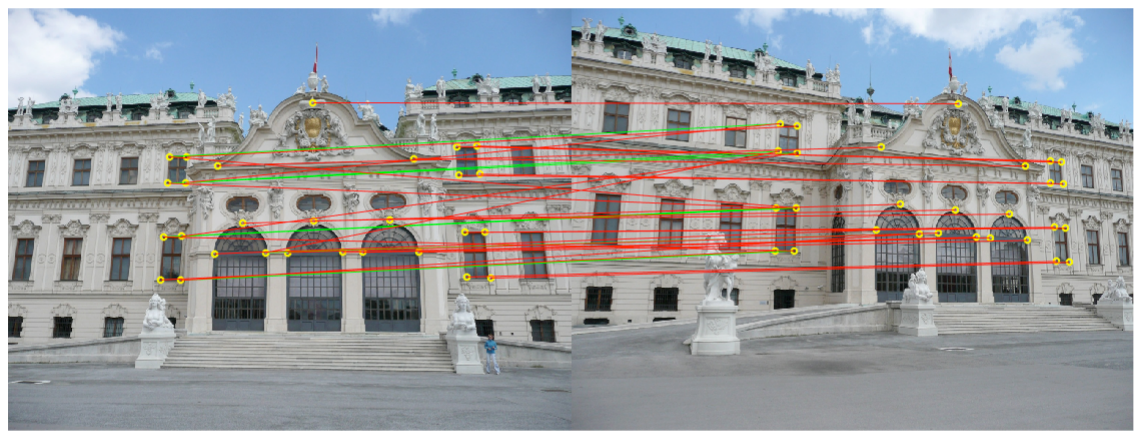}}
    \hspace{.00in}
      \vspace{-.127in}
  \subfigure[Algorithm \ref{alg:improved} with Prim's Order under the Alignment setting]{
    \label{fig:subfig:b} 
    \includegraphics[trim = 8mm 95mm 8mm 83mm, width=0.316\textwidth]{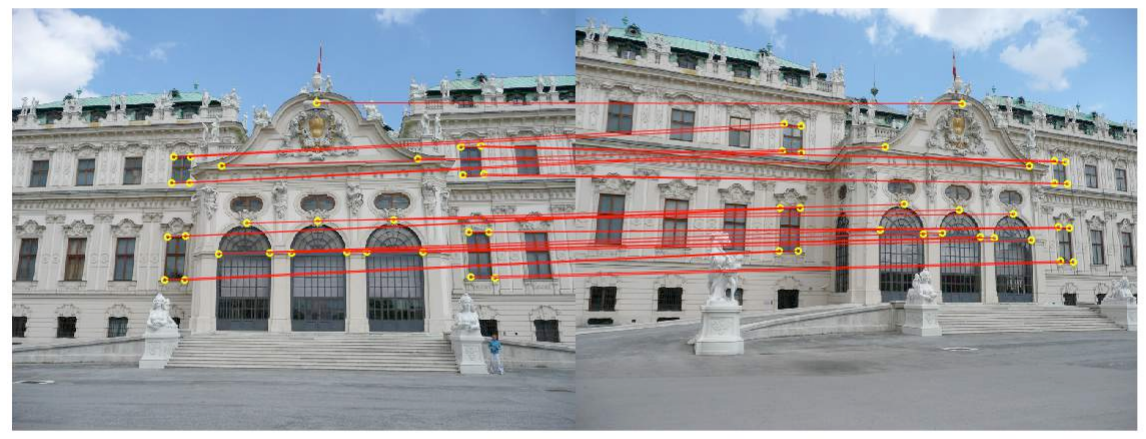}}
  \label{fig:rbf-prim} 
  \hspace{.00in}
    \vspace{-.127in}
 \subfigure[Algorithm \ref{alg:improved} with Kruskal's Order under the Alignment setting]{
    \label{fig:subfig:b} 
    \includegraphics[trim = 8mm 95mm 8mm 83mm, width=0.316\textwidth]{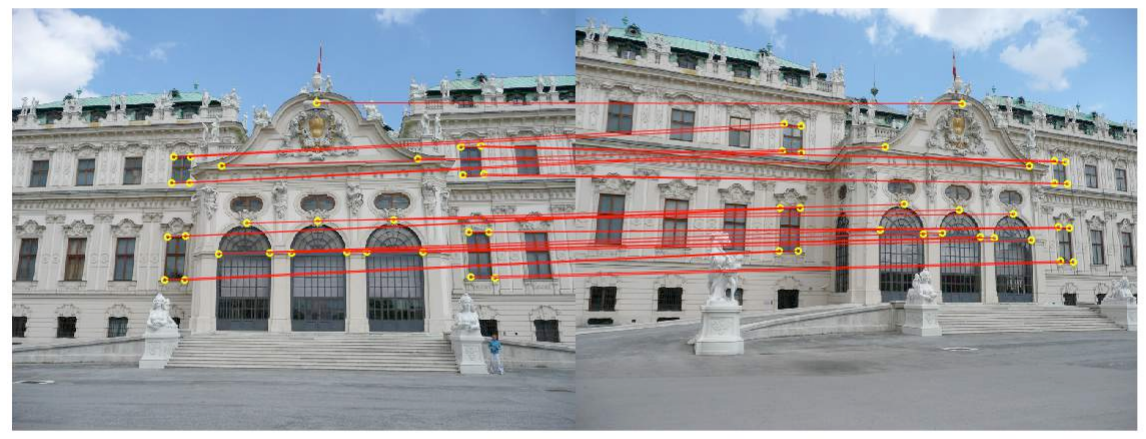}}
  \label{fig:rbf-prim} 
\vspace{.28in}
\caption{Key Points Alignment for the Building dataset. Green lines are the ground truth alignments while red lines are the computed alignments. Less green lines being exposed means a better performance.}
\label{fig:building}
\end{figure*}

Beyond stereo matching, we are also interested in matching images with complicated geometric ambiguities. For instance, images with frequent repetitive structures are extremely hard to handle if we use high-dimensional features such as SIFT \citep{pachauri2013solving}. In this experiment, we use the \textbf{Building} dataset \citep{roberts2011structure} which has such kind of structures. \citet{pachauri2013solving} used this dataset and hand-annotated 25 similar-looking landmark points in the scene across the dataset. We use their hand-annotated data and select $n=14$ images for evaluation. However, in each image, we have $m=28$ key points, and we do not always have all of the 25 landmark points in each scene. Hence, there are many useless key points in each image, which makes the multi-way matching task even harder.

As with the Stereo Landmark Alignments test in \S \ref{sec:expr-CMU}, we explore the RBF setting and the Alignment setting for this experiment. The performance of our Algorithm \ref{alg:improved} is compared against the baseline Permutation Synchronization method as shown in Table \ref{tab:aer}. This task is much more difficult than stereo matching and we do not expect a very high accuracy. Nevertheless, both our methods outperform the baseline algorithm in both settings (more notably in the RBF setting). Furthermore, the alignments we get for each setting and each algorithm are shown in Figure \ref{fig:building}. All of those 6 groups of alignments are between a same pair of images. We can see that, even when we have highly repetitive structures and significant noise in the datasets, our Algorithm \ref{alg:improved} is still stable and gets satisfactory consistent matchings.

\subsection{Experiments in domains beyond computer vision}
Our work aligns multiple objects that are composed of sets of consistent parts. This setting is not limited to images and computer vision tasks. For example, we can apply our algorithms on natural language processing datasets such as the \textbf{Sentence}\footnote{\href{https://archive.ics.uci.edu/ml/datasets/Sentence+Classification}{https://archive.ics.uci.edu/ml/datasets/\\Sentence+Classification}} dataset. This data-set contains human-labeled sentences from 30 research papers. We tried to consistently align the word frequency vectors of each label from different papers. Table \ref{tab:aer} shows that our methods outperform the baseline method, which leads us to believe that our methods could extend to fields beyond computer vision.

\section{CONCLUSION}
\label{sec:conclusion}
Straightforward coordinate updates for the multi-way matching objective are not reliable and produce poor performance. However, if seeded with a good initialization, coordinate updates will (with high probability) not get stuck at bad local optima. By combining a traditional coordinate ascent algorithm with iterative initializations along the edges of a Maximum Spanning Tree (of the graph with edge weights given by pairwise matching similarity values), we obtain stronger empirical results and stronger theoretical guarantees. We outperform leading baseline methods on various problems in computer vision as well as other domains. In addition, our theoretical analysis shows that we do not require all of the noise parameters to be small to ensure a perfect alignment with high probability. Rather, we only require that the spanning tree (on the noise parameter graph) has a small bottleneck edge weight (along with other mild conditions).

\subsubsection*{Acknowledgements}
Work supported in part by DARPA N66001-15-2-4026, N66001-15-C-4032 and NSF III-1526914, IIS-1451500, CCF-1302269. 

\bibliographystyle{plainnat}
\bibliography{aistats2017}

\newpage
\appendix
\section*{SUPPLEMENTARY MATERIAL}
In this supplementary material, we first prove the $n$-way matching objective is NP-hard to optimize. We then prove the Theorems \ref{thm:no-noise} and \ref{thm:noise} in the main paper. Theorem \ref{thm:no-noise2} can be proved using almost the same techniques as the proof for Theorem \ref{thm:no-noise} and hence those details are omitted.
\section{Proof for the NP-hardness of the objective}
\label{sec:proof-np-hard}
\begin{proof}
We show that optimizing the objective in Equation \eqref{obj:v1} is NP-hard by a polynomial time reduction to the known NP-hard problem \textbf{MAX-CUT} \citep{goemans1995improved}, which is to compute the maximum number of edges between a partition of two set of vertices of a given undirected graph. Mathematically, given an undirected graph $G=(V=\{v_1,\ldots,v_n\}, E)$, we want to find partition $(V_1, V_2)$, which satisfies that $V=V_1\cup V_2$, $V_1\cap V_2=\emptyset$, such that $\text{CUT}(V_1, V_2)=|\{(v_i, v_j)\in E:v_i\in V_1, v_j\in V_2\}|$ is maximized.

Given this instance of the MAX-CUT problem, we construct an instance of the multi-way matching problem as follows. Consider optimizing $n$ bijections $\sigma_1,\ldots,\sigma_n$ with the element set size $m=2$. For any $1\le i,j\le n$ and $p,q\in\{1,2\}$, we construct the similarity matrix entry $[T_{i,j}]_{p,q}$ as
\begin{equation}
[T_{i,j}]_{p,q}=\begin{cases}
1,\quad\quad\text{ if }(v_i,v_j)\in E,\text{ and }p\ne q,\\
0,\quad\quad\text{ otherwise}.
\end{cases}
\end{equation}
We can verify that this setting follows the requirements as in \S \ref{sec:preliminary} of the main paper and this construction takes polynomial amount of time (with respect to the instance size of the MAX-CUT problem). 

Notice that, if we denote the set $V_p=\{v_i\in V: \sigma_i(1)=p\}$ for $p\in\{1,2\}$, then $V=V_1\cup V_2$, $V_1\cap V_2=\emptyset$, and $\mathcal L(\sigma_1,\ldots,\sigma_n)=4\text{CUT}(V_1,V_2)$.

Therefore, to compute the max cut for the graph $G$, we just need to compute $\frac14\max\limits_{\sigma_1,\ldots,\sigma_n}\mathcal L(\sigma_1,\ldots,\sigma_n)$. This finishes a polynomial time reduction from the optimization for Equation \eqref{obj:v1} to the MAX-CUT problem. By the NP-hardness of the MAX-CUT problem \citep{goemans1995improved}, we know that the $n$-way matching objective is NP-hard to optimize. Especially, we also know that this objective is NP-hard even for the special case of $m=2$, and hence it is NP-hard for any fixed $m\ge 2$ (by a simple polynomial time reduction, omitted here).
\end{proof}

\section{Proof of Theorem \ref{thm:no-noise}}
\label{sec:proof1}
\begin{proof}
Since we have mentioned that, under the case of this theorem the matrices $P_{ij}$ satisfy the sum $\sum\limits_{i=1}^n\sum\limits_{j=1}^n\text{tr}(P_{ij}^\top T_{ij})$ reaches the optimal value for the objective in Equation \eqref{obj:v2} in the main paper, it is sufficient to show that the matrices $A_1,\ldots,A_n$ returned by Algorithm \ref{alg:original} satisfy $P_{ij}=A_i^\top A_j$ for each $P_{ij}$. We first show that, before the coordinate update part (line 9 to 12) of Algorithm \ref{alg:original}, we have already ensured that the matrices  $A_1,\ldots,A_n$ satisfy the property $P_{ij}=A_i^\top A_j$ for each $P_{ij}$. We will use induction to prove that, after each iteration during the initialization part of the Algorithm \ref{alg:original}, for any set $S_k$ and any $v_i,v_j\in S_k$, we have $P_{ij}=A_i^\top A_j$.
\begin{enumerate}
\item Initially (after the $0^{th}$ iteration), each set $S_k$ only contains one vertex $v_k$. Since $P_{ii}=I=A_k^\top A_k$, the induction assumption is correct.
\item Assume the induction assumption is correct after the $t^{th}$ iteration ($t\ge 0$). For the $(t+1)^{th}$ iteration, denote the edge we use in this iteration as $(v_i, v_j)$. Then, from the algorithm we know that the matrix $\hat P=\argmax\limits_P \text{tr}(P^\top A_iT_{ij}A_j^\top)=A_iP_{ij}A_j^\top$ for the old values of $A_i$ and $A_j$. Therefore, after the update on line 5, we will get $P_{ij}=A_i^\top A_j$ for the new values of $A_i$ and $A_j$. Since we are multiplying on the lefthand side the matrices $A_{j'}$'s on line 5 by the same matrix $\hat P$, this does not break the induction assumption inside the set $S_j$. After the update on line 5, for each $v_{i'}\in S_i$ and each $v_{j'}\in S_j$, we have $A_{i'}^\top A_{j'}=A_{i'}^\top A_{i}A_{i}^\top A_jA_j^\top A_{j'}=P_{i'i}P_{ij}P_{jj'}=P_{i'j'}$. Hence, after line 6 and 7, we know that for each $v_k\in S'$ (the set defined on the line 6) and each $v_{i'}, v_{j'}\in S_k$, we have $P_{i'j'}=A_{i'}^\top A_{j'}$. Since the permutation matrices that are changed during this iteration have their corresponding vertices in the set $S'$, we know that the induction assumption is correct after this iteration.
\end{enumerate}

From 1, 2 we know that we have $P_{ij}=A_i^\top A_j$ for each $P_{ij}$ after initialization. Since we have shown in the main paper that the Pairwise Alignment method can solve the problem optimally on this case, we know that our algorithm has also solved the problem optimally after initialization, and hence we do not have any updates in the coordinate update part. Therefore, Algorithm \ref{alg:original} guarantees an optimal solution in this case.
\end{proof}

\section{Proof of Theorem \ref{thm:noise}}
\label{sec:proof2}
\begin{proof}
Ideally, we want to recover $(A_1,\ldots,A_n)$ such that $A_i^\top A_j=\hat T_{ij}$ for each each pair $(A_i, A_j)$. Let us analyze the probability that we recover such a tuple of $(A_1,\ldots,A_n)$ under the model in Equation \eqref{eqn:model}.

First, let us consider the probability that we recover the correct permutation matrices $\hat T_{ij}$ from the optimization problem $\max\limits_P\text{tr}(P^\top T_{ij})$ for any $i\ne j$. For any permutation matrix $P'\in\mathcal P_m$, $P'\ne T_{ij}$, if we denote $k$ to be the number of entries where $T_{ij}$ equals 1 but $P'$ does not equal 1, then $k=\text{tr}((\hat T_{ij}-P')^\top \hat T_{ij})$. Therefore, $U:=\frac{\text{tr}(P'^\top T_{ij})-\text{tr}(\hat T_{ij}^\top T_{ij})+k}{\eta_{ij}}$ follows the Chi-Square distribution $\chi^2(2k)$. Hence, the probability that $P'$ is a better permutation matrix compared to $\hat T_{ij}$ is
\begin{equation}
\begin{aligned}
&\Pr[\text{tr}(P'^\top T_{ij})\ge \text{tr}(\hat T_{ij}^\top T_{ij})]\\
&=\Pr[\eta_{ij} U-k\ge 0]=\Pr[\frac{U}{\mathbb E[U]}-1\ge\frac12(\frac1{\eta_{ij}}-2)].
\end{aligned}
\end{equation} 

For $\eta_{ij}\le\frac1{10}$, by the Chi-Square tail bounds that \citet{laurent2000adaptive} proposed, 
\begin{equation}
\begin{aligned}
&\Pr[\frac{U_{ij}}{\mathbb E[U_{ij}]}-1\ge\frac12(\frac1{\eta_{ij}}-2)]\\
\le&\Pr[\frac{U_{ij}}{\mathbb E[U_{ij}]}-1\ge\frac14(\frac1{\eta_{ij}}-2)+\sqrt{\frac12(\frac1{\eta_{ij}}-2)}]\\
\le&\exp(-\frac k4(\frac1{\eta_{ij}}-2)).
\end{aligned}
\label{eqn:bound-single-k}
\end{equation}

Denote the probability of misaddressing $k$ letters to $k$ envelopes (The Bernoulli-Euler Problem of the Misaddressed Letters \citep{dorrie2013100}) as $p_k=\sum\limits_{i=0}^\infty\frac{(-1)^i}{i!}\le\frac12$ (for $k\ge 2$). Then, by union bound on Equation \eqref{eqn:bound-single-k} for $k=2,3,\ldots,n$, we know the probability that some $P'\ne\hat T_{ij}$ is better than $\hat T_{ij}$ is at most
\begin{equation}
\begin{aligned}
&\sum\limits_{k=2}^mp_k\cdot\frac{m!}{(m-k)!}\cdot\exp(-\frac k4(\frac1{\eta_{ij}}-2))\\
&\le\frac12\sum\limits_{k=2}^m m^k\cdot\exp(-\frac k4(\frac1{\eta_{ij}}-2)).
\end{aligned}
\end{equation}

If we have $\eta_{ij}\le\frac{1}{4(1+\varepsilon)\ln m+2}$ for some $\varepsilon>0$, then, 
\begin{equation}
\begin{aligned}
&\frac12\sum\limits_{k=2}^m m^k\cdot\exp(-\frac k4(\frac1{\eta_{ij}}-2))\\
&\le\frac12\sum\limits_{k=2}^m m^{-\varepsilon k}=\frac{m^{-2\varepsilon}}{2(1-m^{-\varepsilon})}.
\end{aligned}
\end{equation}

Hence, if we choose the variance parameter $\eta_{ij}\le\min(\frac1{10},\frac{1}{4(1+\varepsilon)\ln m+2})$ for some $\varepsilon>0$, then for $m\ge2^{\frac1\varepsilon}$ we have probability at least $1-m^{-2\varepsilon}$ to guarantee that we recover $\hat T_{ij}$ from the optimization problem $\max\limits_P\text{tr}(P^\top T_{ij})$.

Therefore, if we assume that the number of element sets $n$ is not too large as there exists some constant $\gamma>0$ such that $n\le m^{\gamma}$, then by union bound we know that with probability at least $1-m^{-\delta}$ for any $\delta >0$ that we can guarantee that using the Pairwise Alignment method recovers a correct solution if $m\ge2^{\frac2{4\gamma+\delta}}$ and if we set each $\eta_{ij}\le\min(\frac1{10},\frac{1}{2(2+4\gamma+\delta)\ln m+2})=O(\frac1{\log m})$.

Next let us consider the probability that our Algorithm \ref{alg:original} recovers the correct permutation matrices. 
We would only make some errors on the updates on line 5 and 11. Basically, if we don't make any error at any iteration at the step of computing $\hat P$ on line 4 and don't make any updates on line 11, then we are sure that our algorithm solves the problem optimally.

Here we consider $m\ge 8$ such that $\frac1{10}>\frac{1}{4(1+\varepsilon)\ln m+2}$ for any $\varepsilon>0$. Let us first bound the probability that we might make a mistake when computing the matrix $\hat P$ on line 4. At each iteration when we are considering edge $(v_i, v_j)$, if we have $\eta_{ij}\le\frac{1}{4(1+\varepsilon)\ln m+2}$ for any $\varepsilon>0$, then from the above analysis we know that with probability at least $1-m^{-2\varepsilon}$ we do not make mistakes on this step. 

Otherwise, let us take $(i^*,j^*)=\argmin\limits_{i'\in S_i,j'\in S_j}\eta_{i'j'}$ ($S_i$ and $S_j$ are the sets before being updated on line 7). If we have $\eta_{i^*j^*}\le\frac{1}{8(1+\varepsilon)\ln m+4}\le\frac{1}{4(1+\varepsilon)\ln m+2}$, then from the above analysis we know that with probability at least $1-m^{-2\varepsilon}$ we get $\hat T_{i^*j^*}$ from the optimization problem $\max\limits_{P} \text{tr}(P^\top T_{i^*j^*})$, and we also know that $\eta_{ij}-\eta_{i^*j^*}\ge\frac{1}{8(1+\varepsilon)\ln m+4}$. 

Notice that $(v_i, v_j)$ is an edge of the Maximum Spanning Tree of $G$. It must be the edge with largest edge weight between vertices in $S_i$ and $S_j$. Therefore. we have $f(T_{ij})\ge f(T_{i^*j^*})$. Conditioned on the cases that we recover $\hat T_{i^*j^*}$ from $\max\limits_{P} \text{tr}(P^\top T_{i^*j^*})$ (we will omit some conditional probability notation from now on for brevity), and let $U\sim\chi^2(m)$ be a Chi-Square random variable with free degree $m$, then by the Chi-Square tail bounds that \citet{laurent2000adaptive} proposed,
\begin{equation}
\begin{aligned}
&\Pr[f(T_{i^*j^*})\le m(1-\eta_{i^*j^*}-\frac{1}{16(1+\varepsilon)\ln m+8})]\\
&=\Pr[U-m\ge\frac{m}{\eta_{i^*j^*}(16(1+\varepsilon)\ln m+8)}]\\
&\le\Pr[U-m\ge\frac{m}{2}]\le\Pr[U-m\ge0.48m]\\
&\le\exp(-\frac m{25})\le m^{-2\varepsilon}
\end{aligned}
\label{eqn:Tijstar}
\end{equation}
for sufficiently large $m$. 
On the other hand, consider the value of $f(T_{ij})$, denote $P'=\argmax\limits_{P}\text{tr}(P^\top T_{ij})$ and $k$ to be the number of entries where $\hat T_{ij}$ equals 1 while $P'$ does not equal 1 ($0\le k\le m$). Since we require all $\eta_{ij}\le O(1)$, let us assume that we have $\eta_{ij}\le\frac13$. Let $V_1\sim\chi^2(k)$, $V_2\sim\chi^2(m-k)$ be two independent Chi-Square random variables (we use $\chi^2(0)$ to be the random variable that only has support on a single point 0). Conditioned on $k$, the distribution of $f(T_{ij})$ is the same with $\eta_{ij}(V_1-V_2)+m-k$. If $k>0$, we know that 
\begin{equation}
\begin{aligned}
&\Pr[V_1\ge k+\frac{m}{\eta_{ij}(32(1+\varepsilon)\ln m+16)}]\\
\le&\Pr[V_1- k\ge\frac{3m}{32(1+\varepsilon)\ln m+16}]\\
\le&\Pr[V_1- k\ge2\sqrt{2k\varepsilon\ln m}+4\varepsilon\ln m]\le m^{-2\epsilon}
\end{aligned}
\end{equation}
for sufficiently large $m$. Symmetrically, if $k <  m$,
\begin{equation}
\begin{aligned}
&\Pr[V_2\le (m-k)-\frac{n}{\eta_{ij}(32(1+\varepsilon)\ln m+16)}]\\
\le&\Pr[(m-k)-V_2\ge\frac{3m}{32(1+\varepsilon)\ln m+16}]\\
\le&\Pr[(m-k)-V_2\ge2\sqrt{2k\varepsilon\ln m}]\le m^{-2\epsilon}.
\end{aligned}
\end{equation}
for sufficiently large $m$. Therefore, conditioned on $k$, if we have $\eta_{ij}\le\frac13$, we always have
\begin{equation}
\begin{aligned}
&\Pr[\eta_{ij}(V_1-V_2)+m-k\ge m(1-\eta_{ij}+\frac{1}{16(1+\varepsilon)\ln m+8})]\\
&\le\Pr[\eta_{ij}(V_1-V_2)-(2k-m)\eta_{ij}\ge\frac{1}{16(1+\varepsilon)\ln m+8})]\\
&\le\Pr[V_1\ge k+\frac{m}{\eta_{ij}(32(1+\varepsilon)\ln m+16)}]\\
&+\Pr[V_2\le (m-k)-\frac{m}{\eta_{ij}(32(1+\varepsilon)\ln m+16)}]\le2m^{-2\varepsilon}.
\end{aligned}
\end{equation}

This is true for all $k$. Hence, without conditioning on $k$, we know that
\begin{equation}
\Pr[f(T_{ij})\ge m(1-\eta_{ij}+\frac{1}{16(1+\varepsilon)\ln m+8})]\le 2m^{-2\varepsilon}
\label{eqn:Tij}
\end{equation}
for sufficiently large $m$ and if we have $\eta_{ij}\le\frac13$.

By union bound on Equations \eqref{eqn:Tijstar} and \eqref{eqn:Tij}, we know that, conditioned on the cases where we recover $\hat T_{i^*j^*}$ from $\max\limits_{P} \text{tr}(P^\top T_{i^*j^*})$, since $\eta_{ij}-\eta_{i^*j^*}\ge\frac{1}{8(1+\varepsilon)\ln m+4}$, we have
\begin{equation}
\begin{aligned}
&\Pr[f(T_{ij})\ge f(T_{i^*j^*})]\\
\le&\Pr[f(T_{i^*j^*})\le m(1-\eta_{i^*j^*}-\frac{1}{16(1+\varepsilon)\ln m+8})]\\
 &+\Pr[f(T_{ij})\ge m(1-\eta_{ij}+\frac{1}{16(1+\varepsilon)\ln m+8})]\\
 \le&3m^{-2\varepsilon}.
\end{aligned}
\end{equation}

Since we know that, if we have $\eta_{i^*j^*}\le\frac{1}{8(1+\varepsilon)\ln m+4}$, then with probability at least $1-m^{-2\varepsilon}$ we would recover $\hat T_{i^*j^*}$ from $\max\limits_{P} \text{tr}(P^\top T_{i^*j^*})$. Hence, conditioned on the case that $\eta_{ij}>\frac{1}{4(1+\varepsilon)\ln m+2}$, we know that the probability $\Pr[f(T_{ij})\ge f(T_{i^*j^*})]\le 3m^{-2\varepsilon}+m^{-2\varepsilon}=4m^{-2\varepsilon}$. Plus the opposite case where $\eta_{ij}\le\frac{1}{4(1+\varepsilon)\ln m+2}$, by union bound we know that the probability that we make an error during each iteration of the initialization part of Algorithm \ref{alg:original} is at most $5m^{-2\varepsilon}$. This is true under the condition that $\eta_{ij}\le\frac13$ and $\min\limits_{i'\in S_i,j'\in S_j}\eta_{ij}\le \frac{1}{8(1+\varepsilon)\ln m+4}$. To make these two conditions true, we impose the following two requirements:

\begin{itemize}
\item Consider an undirected weighted graph $G'=(V',E'')$, where there is a vertex $v'_i$ for each element set $X_i$ and their is en edge $(v'_i,v'_j)\in E''$ with edge weight $\eta_{ij}$. Then the bottleneck weight of the minimum bottleneck spanning tree of $G'$ should be at most $\frac{1}{8(1+\varepsilon)\ln m+4}$.
\item $\max\limits_{i<j}\eta_{ij}\le\frac13$.
\end{itemize}

Therefore, under the above two conditions, assume the number of element sets $m$ satisfy $n\le m^\gamma$ for some constant $\gamma>0$. Then, by union bound we know that, for sufficiently large $n$, the probability that we recover the correct solution for $(\hat A_1,\ldots, \hat A_n)$ during the initialization part of the Algorithm \ref{alg:original} is $1-5m^{-2\varepsilon+\gamma}$.

For the coordinate update part of Algorithm \ref{alg:original} (line 9 to 12), let us consider the probability that we do not perform any updates conditioned on the case that we already have an optimal solution in the initialization part. For each step, denote the matrix we are optimizing as $A_i$. The update rule is Equation \eqref{eqn:coor-update}. Using the same approach as before, assume that there is some matrix $P'\ne A_i$ such that $\text{tr}(P'^\top \sum\limits_{1\le j\le n, i\ne j} A_jT_{ij})\ge\text{tr}( A_i^\top \sum\limits_{1\le j\le n, i\ne j} A_jT_{ij})$. Denote $k$ as the number of entries where $ A_i$ equals 1 but $P'$ does not. Also, denote $U_1,\ldots,U_{i-1}$, $U_{i+1},\ldots,U_n$ to be independent random variables following the distribution $\chi^2(2k)$, and let $U$ be a random variable following distribution $\chi^2(2k(n-1))$. Then, by the Chi-Square tail bounds that \citet{laurent2000adaptive} proposed
\begin{equation}
\begin{aligned}
&\Pr[\text{tr}(P'^\top \sum\limits_{1\le j\le n, i\ne j} A_jT_{ij})\ge \text{tr}( A_i^\top \sum\limits_{1\le j\le n, i\ne j} A_jT_{ij})]\\
=&\Pr[\sum\limits_{1\le j\le n,i\ne j}\eta_{ij}U_j\ge k(n-1)]\\
\le&\Pr[\frac13 U\ge k(n-1)]\le\exp(-\frac{2k(n-1)}{25}).
\end{aligned}
\end{equation}

Then, again by union bound on all values for $k$, we know that the probability that we might get a wrong answer for $A_i$ in a single step is at most
\begin{equation}
\begin{aligned}
&\sum\limits_{k=2}^mp_k\cdot\frac{m!}{(m-k)!}\cdot\exp(-\frac{2k(n-1)}{25})\\
\le&\frac12\sum\limits_{k=2}^mm^k\cdot\exp(-\frac{2k(n-1)}{25}).
\end{aligned}
\end{equation}

If we have $n\ge20\ln m$, then for sufficient large value of $m$ we have 
\begin{equation}
\frac12\sum\limits_{k=2}^mm^k\cdot\exp(-\frac{2k(n-1)}{25})\le m^{-2}.
\end{equation}

By union bound on all $m$ matrices $A_i$'s, we know that the probability at least one of them needs updates is at most $m^{-1}$. Hence, we can solve the optimization problem with probability at least $1-5m^{-2\varepsilon+\gamma}+m^{-1}$ under all of the above constraints. If we set $\varepsilon=\frac{\gamma+1}2$, then the probability becomes $1-6m^{-1}=1-o(1)$ for sufficiently large $m$. Under that setting, we require the bottleneck weight of the minimum bottleneck spanning tree of $G'$ to be at at most $\frac{1}{8(1+\varepsilon)\ln m+4}=\frac{1}{4(3+\gamma)\ln m+4}$.
\end{proof}

\end{document}